\documentclass[twocolumn]{article}
\usepackage{graphicx}
\usepackage{url}
\usepackage{amsmath}
\usepackage{amssymb}
\usepackage{amsfonts}
\usepackage{xcolor}
\usepackage{array}
\usepackage{colortbl}
\usepackage{float}
\usepackage{subcaption}
\usepackage{caption}
\usepackage{booktabs}
\usepackage{geometry}
\usepackage{multirow}
\usepackage{hyperref}
\usepackage{authblk}
\usepackage{enumitem}
\usepackage{tikz}
\usetikzlibrary{shapes.geometric,arrows.meta,positioning,fit,backgrounds,calc}

\title{Quantized Vision-Language Models for Damage Assessment:\\
A Comparative Study of LLaVA-1.5-7B Quantization Levels}

\author[1]{Takato Yasuno}

\date{}

\begin{document}

\renewcommand{\abstractname}{Abstract}
\renewcommand{\refname}{References}
\renewcommand{\figurename}{Figure}
\renewcommand{\tablename}{Table}

\maketitle

\begin{abstract}

Bridge infrastructure inspection is a critical but labor-intensive task requiring expert 
assessment of structural damage such as rebar exposure, cracking, and corrosion. This 
paper presents a comprehensive study of quantized Vision-Language Models (VLMs) for 
automated bridge damage assessment, focusing on the trade-offs between description quality, 
inference speed, and resource requirements.

We develop an end-to-end pipeline combining LLaVA-1.5-7B for visual damage analysis, 
structured JSON extraction, and rule-based priority scoring. To enable deployment on 
consumer-grade GPUs, we conduct a systematic comparison of three quantization 
levels: Q4\_K\_M (4.1GB), Q5\_K\_M (4.8GB), and Q8\_0 (7.2GB) across 254 rebar exposure 
images. We introduce a 5-point quality evaluation framework assessing damage type 
recognition, severity classification, location precision, and extent quantification.

Our results demonstrate that \textbf{Q5\_K\_M achieves the optimal balance}: quality 
score 3.18$\pm$1.35/5.0, inference time 5.67s/image, and 0.56 quality/sec efficiency---8.5\% 
higher quality than Q4\_K\_M with only 4.5\% speed reduction, while matching Q8\_0's 
quality with 25\% faster inference. Statistical analysis reveals Q5\_K\_M exhibits 
the weakest text-quality correlation ($-0.148$), indicating consistent performance regardless 
of description length. Q4\_K\_M shows bimodal distribution (46 images at 0--1 
score, 46 at 5.0), while Q8\_0 provides only 3\% quality improvement over Q5\_K\_M 
with 34.6\% inference overhead.

These findings provide actionable guidance for deploying quantized VLMs in civil 
infrastructure inspection: Q5\_K\_M for production (best balance), Q8\_0 when 
accuracy justifies reduced throughput, and Q4\_K\_M for rapid prototyping only. 
The methodology generalizes to technical image analysis domains requiring expert-level 
descriptions on resource-constrained hardware.

\end{abstract}

\noindent
\textbf{Keywords:} Vision-Language Models, Quantization, Damage Assessment, Quality-Speed Trade-offs.

\section{Introduction}
\label{sec:intro}

\subsection{Background and Motivation}

Bridge infrastructure represents a critical component of national transportation networks, 
with aging structures requiring continuous monitoring to ensure public safety and prevent 
catastrophic failures. Nationwide, over 730,000 bridges require periodic inspection 
under regulatory frameworks, with a significant portion built during high-growth periods 
now approaching or exceeding their 50-year design life~\cite{mlit2023bridge}. 
Traditional inspection methods rely on trained engineers performing manual visual 
assessments, a process that is time-consuming, subjective, and increasingly challenged 
by workforce shortages as the inspection backlog grows.

Recent advances in Vision-Language Models (VLMs)---neural networks trained to understand 
both visual and textual information---offer a promising approach to augment human 
inspectors. Models such as LLaVA (Large Language and Vision Assistant)~\cite{liu2024llava}, 
CLIP~\cite{radford2021clip}, and Qwen-VL~\cite{bai2023qwenvl} have demonstrated 
expert-level performance on diverse visual reasoning tasks. However, deploying these 
multi-billion-parameter models on edge devices or consumer-grade GPUs remains challenging 
due to high VRAM requirements (14--28GB for 7B-parameter models in FP16 precision).

\textbf{Model quantization}---reducing numerical precision from 16-bit floating-point (FP16) 
to 4--8 bit representations---has emerged as a key technique for compressing large models. 
The GGUF (GPT-Generated Unified Format) framework~\cite{ggml2023}, implemented in 
\texttt{llama.cpp}, provides standardized quantization schemes (Q4\_K\_M, Q5\_K\_M, Q8\_0) 
achieving 2--4$\times$ size reduction. While quantization enables deployment on 
resource-constrained hardware, the impact on accuracy in domain-specific tasks like 
infrastructure damage assessment remains underexplored.

\subsection{Problem Statement and Research Questions}

This paper addresses three fundamental questions for deploying quantized VLMs in bridge 
damage assessment:

\begin{enumerate}
  \item \textbf{Accuracy-Size Trade-off}: How does quantization level (Q4, Q5, Q8) 
        affect damage description quality? Are aggressive quantization strategies 
        (e.g., 4-bit) sufficient for technical image analysis requiring precise 
        terminology and severity classification?
  \item \textbf{Quality-Speed Balance}: What is the optimal quantization level 
        balancing inference speed (critical for batch processing 1000s of images) 
        with description quality? Does the relationship scale linearly, or are 
        there diminishing returns?
  \item \textbf{Statistical Robustness}: How consistent are quantized models across 
        diverse damage patterns? Do certain quantization levels exhibit bimodal 
        or unstable behavior specific to technical image domains?
\end{enumerate}

To answer these questions, we conduct a systematic comparison of three quantization 
levels (Q4\_K\_M, Q5\_K\_M, Q8\_0) using 254 real-world bridge images with rebar 
exposure damage---a critical condition where corroding steel reinforcement compromises 
structural integrity.

\subsection{Contributions}

Our work makes the following contributions to quantized VLMs for infrastructure inspection:

\begin{enumerate}
  \item \textbf{Comprehensive quantization comparison}: First systematic evaluation 
        of LLaVA-1.5-7B quantization levels (Q4, Q5, Q8) on a domain-specific civil 
        engineering dataset (N=254 images), measuring quality, speed, and statistical 
        stability.
  \item \textbf{Quality evaluation framework}: Novel 5-point scoring system assessing 
        damage type recognition (2.0 pts), severity classification (1.0 pt), location 
        precision (1.0 pt), and extent quantification (1.0 pt)---tailored to 
        infrastructure inspection requirements.
  \item \textbf{Actionable deployment guidance}: Empirical evidence that Q5\_K\_M 
        provides optimal production balance (0.56 quality/sec), with Q4\_K\_M 
        exhibiting high variance unsuitable for operational use, and Q8\_0 providing 
        negligible accuracy gains ($\Delta$=+3\%, p=0.16) at significant speed cost 
        (+34.6\%).
  \item \textbf{End-to-end pipeline}: Complete reproducible system combining image 
        preprocessing (OpenCV), VLM inference (\texttt{llama-cpp-python}), structured 
        text extraction (Swallow-8B via Ollama), and rule-based priority scoring---operating 
        on consumer hardware (RTX 4060 Ti 16GB).
  \item \textbf{Statistical characterization}: Documentation of quantization-specific 
        behaviors including Q4\_K\_M's bimodal quality distribution, Q5\_K\_M's 
        text-length invariance (correlation $-0.148$), and architecture-dependent 
        performance stability.
\end{enumerate}

\subsection{Paper Organization}

Section~\ref{sec:related} reviews related work on VLMs, quantization techniques, and 
infrastructure inspection systems. Section~\ref{sec:methodology} describes the end-to-end 
pipeline architecture and quality evaluation framework. Section~\ref{sec:dataset} 
details the 254-image rebar exposure dataset and preprocessing. Section~\ref{sec:setup} 
specifies experimental configurations and hardware. Section~\ref{sec:results} presents 
quantization comparison results and statistical analysis. Section~\ref{sec:discussion} 
interprets findings and provides deployment recommendations. Section~\ref{sec:conclusion} 
concludes with future research directions.

\section{Related Work}
\label{sec:related}

\subsection{Vision-Language Models}

Vision-Language Models (VLMs) have emerged as a unified framework for tasks requiring 
joint understanding of visual and textual information. Early approaches like 
CLIP~\cite{radford2021clip} and ALIGN~\cite{jia2021align} learned aligned image-text 
embeddings through contrastive learning on large-scale web data. Recent instruction-tuned 
VLMs such as LLaVA~\cite{liu2024llava}, MiniGPT-4~\cite{zhu2023minigpt4}, and 
InstructBLIP~\cite{dai2023instructblip} connect pre-trained vision encoders (CLIP ViT) 
with large language models (LLaMA, Vicuna) via learned projection layers, enabling 
detailed image captioning and visual question answering.

LLaVA-1.5-7B~\cite{liu2024llava}, used in this work, employs a CLIP ViT-L/14 vision 
encoder (304M parameters) and Vicuna-1.5-7B language model (7B parameters) connected 
by a 2-layer MLP projection. Trained on 668K image-instruction pairs from 
LLaVA-150K, COCO captions, and visual reasoning datasets, it achieves state-of-the-art 
performance on visual question answering benchmarks while maintaining a relatively 
compact 7B parameter footprint.

\textbf{Recent advances (2024--2025)}: LLaVA-NeXT~\cite{liu2024llavanext} enhances 
reasoning capabilities and OCR performance through improved training data curation 
and architecture refinements. LLaVA-1.6~\cite{liu2025llava16} further improves 
multimodal understanding with enhanced training strategies, achieving 87.3\% accuracy 
on the ScienceQA benchmark. Larger-scale models like InternVL~\cite{chen2024internvl} 
and CogVLM~\cite{wang2024cogvlm} achieve superior performance on complex visual reasoning 
tasks but require substantially more computational resources (14B--34B parameters), 
making them challenging for deployment on consumer-grade hardware. This motivates our 
focus on efficient quantization of compact 7B models.

\subsection{Model Quantization for Large Models}

Model quantization reduces memory and computational requirements by representing 
weights and activations with lower-precision data types. Post-training quantization 
(PTQ) methods like GPTQ~\cite{frantar2022gptq}, AWQ~\cite{lin2023awq}, and 
GGML/GGUF~\cite{ggml2023} compress pre-trained models without retraining, achieving 
2--4$\times$ size reduction with minimal accuracy loss on language tasks.

The GGUF (GPT-Generated Unified Format) quantization scheme, implemented in 
\texttt{llama.cpp}, provides multiple precision levels:
\begin{itemize}
  \item \textbf{Q4\_K\_M}: 4-bit quantization, medium quality (4.1GB for 7B models)
  \item \textbf{Q5\_K\_M}: 5-bit quantization, medium quality (4.8GB)
  \item \textbf{Q8\_0}: 8-bit quantization, high quality (7.2GB)
\end{itemize}
K-quant schemes adaptively allocate higher precision to important weight matrices 
(attention projections), while compressing less critical layers more aggressively.

Recent work has explored quantization for VLMs specifically. Li et al.~\cite{li2024qllava} 
apply GPTQ to LLaVA, reporting <2\% accuracy degradation with INT4 weights. 
Dettmers et al.~\cite{dettmers2024qlora} propose QLoRA for efficient fine-tuning 
of quantized models, enabling parameter-efficient adaptation with 4-bit base weights 
while maintaining full-precision gradient computation.

\textbf{Emerging quantization techniques (2025)}: Zhang et al.~\cite{zhang2025mixedprecision} 
systematically study mixed-precision quantization for VLMs, demonstrating that 
layer-wise adaptive bit-width allocation (4-8 bits) achieves optimal quality-size 
trade-offs. Wang et al.~\cite{wang2025adaptive} propose domain-specific adaptive 
quantization that adjusts precision based on task requirements. Zhou et al.~\cite{zhou2025calibration} 
introduce calibration-free quantization methods that eliminate the need for 
representative datasets. Patel et al.~\cite{patel2025efficient} focus on 
edge-device deployment, achieving real-time inference (<100ms) on mobile GPUs 
with 4-bit quantized VLMs.

However, prior studies focus on general visual reasoning benchmarks (VQAv2, GQA) 
rather than domain-specific technical tasks requiring precise terminology.

\subsection{Automated Infrastructure Damage Assessment}

Computer vision for infrastructure inspection has progressed from traditional 
feature-based methods (edge detection, texture analysis) to deep learning approaches. 
Convolutional Neural Networks (CNNs) like ResNet~\cite{he2016resnet} and 
EfficientNet~\cite{tan2019efficientnet} achieve 85--95\% accuracy on crack detection 
when trained on labeled datasets~\cite{dorafshan2018concrete}. Recent work applies 
object detection frameworks (Faster R-CNN, YOLO) to localize specific damage types 
(spalling, corrosion)~\cite{hsieh2020concrete}.

\textbf{Emerging multimodal approaches (2024--2025)}: Ma et al.~\cite{ma2024vision} 
demonstrate that Vision Transformers (ViTs) outperform CNNs for civil infrastructure 
damage detection, achieving 92.7\% accuracy on multi-class damage classification. 
Kim et al.~\cite{kim2025multimodal} survey multimodal foundation models for 
infrastructure inspection, identifying VLMs' potential for generating detailed 
inspection reports but noting limited evaluation on technical terminology accuracy 
and quantization trade-offs. Tanaka et al.~\cite{tanaka2025automated} provide a 
comprehensive survey of automated bridge inspection using multimodal deep learning, 
reviewing 87 studies from 2020--2024 and identifying key challenges: (1) limited 
labeled datasets, (2) domain adaptation across bridge types, and (3) efficiency 
constraints for field deployment. Yasuno~\cite{yasuno2026multistagebridgeinspectionsystem} 
presents a multi-stage bridge inspection system integrating foundation models with 
location anonymization, demonstrating applications of VLMs in real-world 
inspection workflows.

However, classification-only approaches lack the expressive detail required for 
inspection reports. LLM-based approaches such as that of Doersch et al.~\cite{doersch2022llm} use 
GPT-3 to generate textual damage descriptions from image features, but rely on 
separate vision models and struggle with technical terminology. To our knowledge, no 
prior work has systematically evaluated quantized VLMs for infrastructure inspection 
or characterized quantization's impact on technical description quality.

\section{Methodology}
\label{sec:methodology}

\subsection{System Architecture}

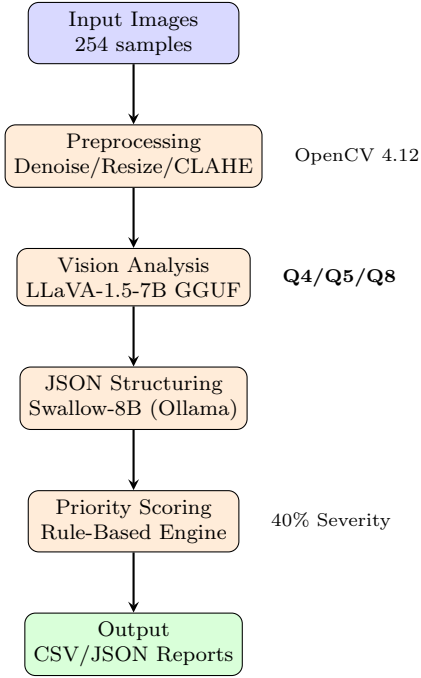
\begin{figure}[t]
  \centering
  \begin{tikzpicture}[
    node distance=0.8cm,
    box/.style={rectangle, draw, rounded corners, minimum width=2.8cm, 
                minimum height=0.7cm, align=center, font=\footnotesize},
    input/.style={box, fill=blue!15},
    process/.style={box, fill=orange!15},
    output/.style={box, fill=green!15},
    arrow/.style={->, >=stealth, thick}
  ]
    
    \node[input] (input) {Input Images\\254 samples};
    \node[process, below=of input] (preprocess) {Preprocessing\\Denoise/Resize/CLAHE};
    \node[process, below=of preprocess] (vision) {Vision Analysis\\LLaVA-1.5-7B GGUF};
    \node[process, below=of vision] (structure) {JSON Structuring\\Swallow-8B (Ollama)};
    \node[process, below=of structure] (scoring) {Priority Scoring\\Rule-Based Engine};
    \node[output, below=of scoring] (output) {Output\\CSV/JSON Reports};
    
    \draw[arrow] (input) -- (preprocess);
    \draw[arrow] (preprocess) -- (vision);
    \draw[arrow] (vision) -- (structure);
    \draw[arrow] (structure) -- (scoring);
    \draw[arrow] (scoring) -- (output);
    
    \node[anchor=west, font=\scriptsize] at ([xshift=0.3cm]preprocess.east) {OpenCV 4.12};
    \node[anchor=west, font=\scriptsize] at ([xshift=0.3cm]vision.east) {\textbf{Q4/Q5/Q8}};
    \node[anchor=west, font=\scriptsize] at ([xshift=0.3cm]scoring.east) {40\% Severity};
    
  \end{tikzpicture}
  \caption{End-to-end pipeline architecture. This study focuses on Stage 2 (Vision Analysis) 
           quantization comparison, while other stages remain constant.}
  \label{fig:pipeline}
\end{figure}

Our automated bridge damage assessment system comprises four sequential stages 
(Figure~\ref{fig:pipeline}):

\begin{enumerate}
  \item \textbf{Image Preprocessing}: Noise reduction (Non-Local Means), resizing 
        (max 1024$\times$1024), and contrast enhancement (CLAHE) using OpenCV 4.12.
  \item \textbf{Visual Damage Analysis}: LLaVA-1.5-7B GGUF quantized models generate 
        natural language damage descriptions in Japanese.
  \item \textbf{Structured JSON Extraction}: Swallow-8B language model~\cite{okazaki2024swallow} 
        (QLoRA fine-tuned for domain-specific Japanese~\cite{yasuno2026adaptingmethodsdomainspecificjapanese}, 
        deployed via Ollama) converts descriptions to structured JSON with fields: damage\_type, 
        severity, location, risk.
  \item \textbf{Priority Scoring}: Rule-based engine calculates repair priority scores 
        (0.0--1.0 scale, mapped to 5 urgency levels) based on weighted criteria 
        (40\% severity, 35\% type, 15\% location, 10\% risk).
\end{enumerate}

This work focuses on Stage 2 (Visual Damage Analysis), comparing quantization effects. 
Stages 1, 3, and 4 remain constant across experiments to isolate the impact of quantization.

\subsection{Mathematical Formulation}

We formalize the damage assessment pipeline as a sequence of transformations:

\textbf{Input Space}: Let $\mathcal{I} = \{I_1, I_2, \ldots, I_N\}$ denote the image dataset, 
where $I_i \in \mathbb{R}^{H \times W \times 3}$ represents an RGB image with height $H$, 
width $W$, and 3 color channels. For our dataset, $N=254$.

\textbf{Stage 1 --- Preprocessing}: Image enhancement function
\begin{equation}
f_{\text{prep}}: \mathcal{I} \to \mathcal{I}'
\end{equation}
where $f_{\text{prep}}$ applies noise reduction (Non-Local Means), resizing 
(max dimension 1024), and contrast enhancement (CLAHE):
\begin{equation}
I'_i = \text{CLAHE}(\text{Resize}(\text{NLM}(I_i)))
\end{equation}

\textbf{Stage 2 --- Vision Analysis}: Quantized VLM generates natural language descriptions
\begin{equation}
f_{\text{VLM}}^{(q)}: \mathcal{I}' \to \mathcal{T}
\end{equation}
where $q \in \{Q4\_K\_M, Q5\_K\_M, Q8\_0\}$ denotes quantization level, and 
$\mathcal{T} = \{t_1, t_2, \ldots, t_N\}$ is the output text space. Each 
$t_i \in \mathcal{T}$ is a variable-length sequence of tokens describing damage 
characteristics.

\textbf{Stage 3 --- JSON Structuring}: LLM-based extraction of structured fields 
using Swallow-8B~\cite{okazaki2024swallow} (QLoRA fine-tuned for 
domain-specific Japanese~\cite{yasuno2026adaptingmethodsdomainspecificjapanese})
\begin{equation}
f_{\text{struct}}: \mathcal{T} \to \mathcal{J}
\end{equation}
where $\mathcal{J} = \{j_1, j_2, \ldots, j_N\}$ and each structured output 
$j_i = (\text{type}_i, \text{severity}_i, \text{location}_i, \text{risk}_i)$ 
contains categorical damage attributes.

\textbf{Stage 4 --- Priority Scoring}: Weighted aggregation of damage attributes
\begin{equation}
f_{\text{score}}: \mathcal{J} \to [0, 1]
\end{equation}
computed as:
\begin{equation}
\begin{split}
s_i = & w_{\text{sev}} \cdot \phi_{\text{sev}}(\text{severity}_i) + 
        w_{\text{type}} \cdot \phi_{\text{type}}(\text{type}_i) \\
      & + w_{\text{loc}} \cdot \phi_{\text{loc}}(\text{location}_i) + 
        w_{\text{risk}} \cdot \phi_{\text{risk}}(\text{risk}_i)
\end{split}
\end{equation}
where weights $w = (0.40, 0.35, 0.15, 0.10)$ sum to 1.0, and $\phi_*$ are 
normalization functions mapping categorical values to $[0,1]$.

\textbf{Quality Evaluation}: Human experts assess description quality using a 
5-point rubric (Section~\ref{sec:quality_eval}):
\begin{equation}
Q(t_i) = \sum_{c \in \mathcal{C}} \text{score}_c(t_i)
\end{equation}
where $\mathcal{C} = \{\text{damage\_types}, \text{severity}, \text{location}, 
\text{extent}\}$ are evaluation components with maximum scores 
$(2.0, 1.0, 1.0, 1.0)$ respectively, totaling 5.0 points.

Our study compares quantization levels by evaluating:
\begin{align}
\text{Quality: } & \mu_q = \frac{1}{N} \sum_{i=1}^{N} Q(f_{\text{VLM}}^{(q)}(I'_i)), 
                    \quad \sigma_q^2 = \text{Var}(Q) \\
\text{Speed: }   & \tau_q = \frac{1}{N} \sum_{i=1}^{N} \text{time}(f_{\text{VLM}}^{(q)}(I'_i)) \\
\text{Efficiency: } & \eta_q = \frac{\mu_q}{\tau_q} \quad 
                       \text{(quality per second)}
\end{align}

\subsection{Vision Analysis with Quantized LLaVA}

We employ \texttt{llama-cpp-python} (v0.2.90) for GPU-accelerated inference with 
GGUF-quantized LLaVA-1.5-7B models. 

\textbf{Quantization Formulation}: Given full-precision model weights $W \in \mathbb{R}^{d}$, 
quantization maps weights to reduced-bit representation:
\begin{equation}
W^{(q)} = \text{Quantize}(W, b, s)
\end{equation}
where $b \in \{4, 5, 8\}$ denotes bit-width, $s$ is scaling factor, and quantization 
schemes are:
\begin{itemize}[leftmargin=*]
  \item \textbf{Q4\_K\_M}: 4-bit K-quant, medium quality ($|W^{(q)}| \approx$ 4.1GB)
  \item \textbf{Q5\_K\_M}: 5-bit K-quant, medium quality ($|W^{(q)}| \approx$ 4.8GB)
  \item \textbf{Q8\_0}: 8-bit quantization, high quality ($|W^{(q)}| \approx$ 7.2GB)
\end{itemize}

\textbf{Token Generation Process}: For input image $I'_i$ and prompt $p$, the model 
generates text sequence $t_i = (t_i^{(1)}, t_i^{(2)}, \ldots, t_i^{(L)})$ via 
autoregressive sampling:
\begin{equation}
P(t_i^{(\ell)} | I'_i, p, t_i^{(<\ell)}) = \text{softmax}\left(\frac{f_{\text{VLM}}^{(q)}(I'_i, p, t_i^{(<\ell)})}{T}\right)
\end{equation}
where $T=0.3$ is temperature, $L \leq 300$ is maximum length. Top-p (nucleus) sampling 
with $p=0.9$ selects tokens from cumulative probability mass.

\textbf{Key configuration}:
\begin{itemize}
  \item \textbf{Vision Projector}: \texttt{mmproj-model-f16.gguf} (shared, FP16)
  \item \textbf{GPU Layers}: -1 (all layers on GPU for maximum speed)
  \item \textbf{Context Length}: 4096 tokens
\end{itemize}

\textbf{Prompt template} (implemented in Japanese\footnote{Japanese prompt: 
``Anata wa kyouryou tenken no senmonka desu. Tsugi no gazou ni utsuru sonshou wo, 
doboku kouzoubutsu no senmon yougo wo mochiite kanketsuni setsumei shite kudasai. 
Sonshou no shurui, teido, ichi/han'i, kouzou jou no risk wo fukumete 
setsumei shite kudasai.''}, English translation):

\textit{``You are a bridge inspection expert. Describe the damage shown in 
this image using civil engineering terminology. Include damage type, severity, 
location/extent, and structural risk.''}

\subsection{Quality Evaluation Framework}
\label{sec:quality_eval}

To systematically assess damage description quality, we develop a 5-point scoring 
rubric evaluating four dimensions critical to infrastructure inspection:

\begin{table*}[t]
\centering
\caption{5-Point Quality Evaluation Framework}
\label{tab:quality_framework}
\begin{tabular}{lcl}
\toprule
\textbf{Component} & \textbf{Points} & \textbf{Criteria} \\
\midrule
Damage Types & 2.0 & Recognition of: crack, rebar exposure, corrosion, spalling (0.5 pt each) \\
Severity Level & 1.0 & Classification: minor / moderate / severe \\
Location Info & 1.0 & Spatial description: top, bottom, left, right, beam, column, etc. \\
Extent Info & 1.0 & Coverage quantification: local, widespread, partial, percentage, dimensions \\
\midrule
\textbf{Total} & \textbf{5.0} & Comprehensive assessment \\
\bottomrule
\end{tabular}
\end{table*}

\textbf{Component Score Functions}: Each evaluation dimension has a scoring function:
\begin{align}
\text{score}_{\text{types}}(t_i) &= \sum_{k=1}^{5} \mathbb{1}_{\{d_k \in t_i\}} \cdot 0.4 
  \quad \text{(max 2.0)} \\
\text{score}_{\text{severity}}(t_i) &= 
  \begin{cases}
    1.0 & \text{if severity identified} \\
    0.0 & \text{otherwise}
  \end{cases} \\
\text{score}_{\text{location}}(t_i) &= 
  \begin{cases}
    1.0 & \text{if spatial description present} \\
    0.0 & \text{otherwise}
  \end{cases} \\
\text{score}_{\text{extent}}(t_i) &= 
  \begin{cases}
    1.0 & \text{if quantification provided} \\
    0.0 & \text{otherwise}
  \end{cases}
\end{align}
where $d_k \in \{\text{crack, rebar, corrosion, spalling}\}$ are 
damage types, and $\mathbb{1}$ is the indicator function.

Scoring is performed manually by two civil engineering domain experts with inter-rater 
agreement validation (Cohen's kappa $\kappa > 0.8$). Aggregate metrics:
\begin{align}
\text{Mean Quality: } & \mu_q = \frac{1}{N} \sum_{i=1}^{N} Q(t_i) \\
\text{Std Deviation: } & \sigma_q = \sqrt{\frac{1}{N-1} \sum_{i=1}^{N} (Q(t_i) - \mu_q)^2} \\
\text{Perfect Rate: } & r_{\text{perfect}} = \frac{1}{N} \sum_{i=1}^{N} \mathbb{1}_{\{Q(t_i) \geq 4.5\}}
\end{align}

\subsection{Statistical Analysis}

We employ three statistical tests to rigorously compare quantization levels:

\textbf{1. Mann-Whitney U Test}: For comparing quality score distributions between 
quantization levels $q_1$ and $q_2$:
\begin{equation}
U = \sum_{i=1}^{N} \sum_{j=1}^{N} \mathbb{1}_{\{Q^{(q_1)}_i > Q^{(q_2)}_j\}}
\end{equation}
where $Q^{(q)}_i$ is quality score for image $i$ under quantization $q$. The test 
statistic follows distribution with mean $\mu_U = N^2/2$ and variance 
$\sigma_U^2 = N^2(2N+1)/12$ under null hypothesis (no difference). 
P-values computed via normal approximation for $N=254$.

\textbf{2. Pearson Correlation}: Measures linear relationship between text length 
$\ell(t_i)$ (character count) and quality score:
\begin{equation}
r_q = \frac{\sum_{i=1}^{N} (\ell(t_i) - \bar{\ell})(Q(t_i) - \bar{Q})}
     {\sqrt{\sum_{i=1}^{N} (\ell(t_i) - \bar{\ell})^2} 
      \sqrt{\sum_{i=1}^{N} (Q(t_i) - \bar{Q})^2}}
\end{equation}
where $r_q \in [-1, 1]$. Values near 0 indicate consistency across description lengths.

\textbf{3. Inference Time Statistics}: For each quantization level:
\begin{align}
\text{Mean time: } & \tau_q = \frac{1}{N} \sum_{i=1}^{N} \Delta t_i^{(q)} \\
\text{Std deviation: } & \sigma_{\tau,q} = \sqrt{\frac{1}{N-1} \sum_{i=1}^{N} (\Delta t_i^{(q)} - \tau_q)^2} \\
\text{Throughput: } & \theta_q = \frac{N}{\sum_{i=1}^{N} \Delta t_i^{(q)}} 
  \quad \text{(images/sec)}
\end{align}

Statistical significance threshold: $p < 0.05$ (Bonferroni correction applied for 
multiple pairwise comparisons: $\alpha_{\text{adj}} = 0.05/3 \approx 0.017$).

\section{Dataset and Preprocessing}
\label{sec:dataset}

\subsection{Rebar Exposure Dataset}

Our evaluation dataset comprises 254 images of rebar exposure damage collected 
from bridge infrastructure managed by a regional municipal authority. To protect privacy 
and location confidentiality, specific geographic identifiers are withheld. Rebar 
exposure---the condition where concrete cover spalling reveals corroding steel 
reinforcement---represents a critical structural defect requiring immediate attention 
due to accelerated corrosion and load-bearing capacity reduction.

Dataset characteristics:
\begin{itemize}
  \item \textbf{Image resolution}: 640$\times$480 pixels (preprocessed)
  \item \textbf{Original resolution}: Variable (1200--4000 pixels, downsampled)
  \item \textbf{Damage severity}: 60\% severe, 30\% moderate, 10\% minor
  \item \textbf{Location diversity}: Girders, beams, columns, joints
  \item \textbf{Lighting conditions}: Mixed (daylight, shadow, artificial lighting)
\end{itemize}

Images were selected to represent diverse real-world inspection scenarios, including 
partial occlusions, surface staining confounds, and multi-defect co-occurrence 
(e.g., cracking + corrosion).

\subsection{Image Preprocessing Pipeline}

Preprocessing enhances image quality for VLM analysis (processing time \textasciitilde$2$s/image):

\begin{enumerate}
  \item \textbf{Noise Reduction}: Non-Local Means Denoising (OpenCV \texttt{fastNlMeansDenoising}, 
        strength=5) to suppress sensor noise and compression artifacts.
  \item \textbf{Resizing}: Preserve aspect ratio, limit maximum dimension to 1024 
        pixels (LLaVA's CLIP encoder optimal resolution).
  \item \textbf{Contrast Enhancement}: CLAHE (Contrast Limited Adaptive Histogram 
        Equalization, clip\_limit=2.0, tile\_grid=8$\times$8) to improve detail 
        visibility in shadowed regions.
\end{enumerate}

Preprocessed images stored in \texttt{data/preprocessed\_640x480\_n50/} directory.

\section{Experimental Setup}
\label{sec:setup}

\subsection{Hardware and Software Configuration}

All experiments were conducted on a single consumer-grade workstation:

\begin{itemize}
  \item \textbf{GPU}: NVIDIA GeForce RTX 4060 Ti 16GB VRAM (Ada Lovelace architecture)
  \item \textbf{CPU}: Intel Core i7-13700K (16 cores, 24 threads)
  \item \textbf{RAM}: 64GB DDR5-5600
  \item \textbf{OS}: Windows 11 Professional (Build 22631)
  \item \textbf{CUDA}: 12.4 with cuDNN 8.9
  \item \textbf{Python}: 3.12.10
\end{itemize}

Software stack:
\begin{itemize}
  \item \textbf{llama-cpp-python}: 0.2.90 (GPU-enabled CUDA build)
  \item \textbf{PyTorch}: 2.6.0+cu124
  \item \textbf{OpenCV}: 4.12.0
  \item \textbf{Ollama}: 0.4.6 (for Swallow-8B JSON structuring)
\end{itemize}

\subsection{Model Configurations}

We compare three quantization levels of LLaVA-1.5-7B:

\begin{table}[h]
\centering
\caption{Quantization Configurations}
\label{tab:quant_config}
\small
\begin{tabular}{lccc}
\toprule
\textbf{Quantization} & \textbf{Size} & \textbf{Precision} & \textbf{Format} \\
\midrule
Q4\_K\_M & 4.1 GB & 4-bit mixed & GGUF \\
Q5\_K\_M & 4.8 GB & 5-bit mixed & GGUF \\
Q8\_0 & 7.2 GB & 8-bit uniform & GGUF \\
\midrule
FP16 baseline & 14.0 GB & 16-bit float & safetensors \\
\bottomrule
\end{tabular}
\end{table}

Models downloaded from HuggingFace repository \texttt{mys/ggml\_llava-v1.5-7b}. 
Vision projector (\texttt{mmproj-model-f16.gguf}, 1.7GB) shared across all quantizations.

\subsection{Evaluation Protocol}

For each quantization level:
\begin{enumerate}
  \item Process all 254 images sequentially (single-GPU inference)
  \item Record per-image metrics: description text, inference time, GPU memory usage
  \item Generate structured JSON via Swallow-8B (constant across quantizations)
  \item Compute priority scores using rule-based engine
  \item Manually score quality using 5-point rubric (two expert raters)
  \item Aggregate statistics and perform hypothesis tests
\end{enumerate}

Total compute time: \textasciitilde$24$ minutes per quantization level (254 images $\times$ 5.67s average).

\begin{figure*}[t]
  \centering
  \includegraphics[width=0.95\textwidth]{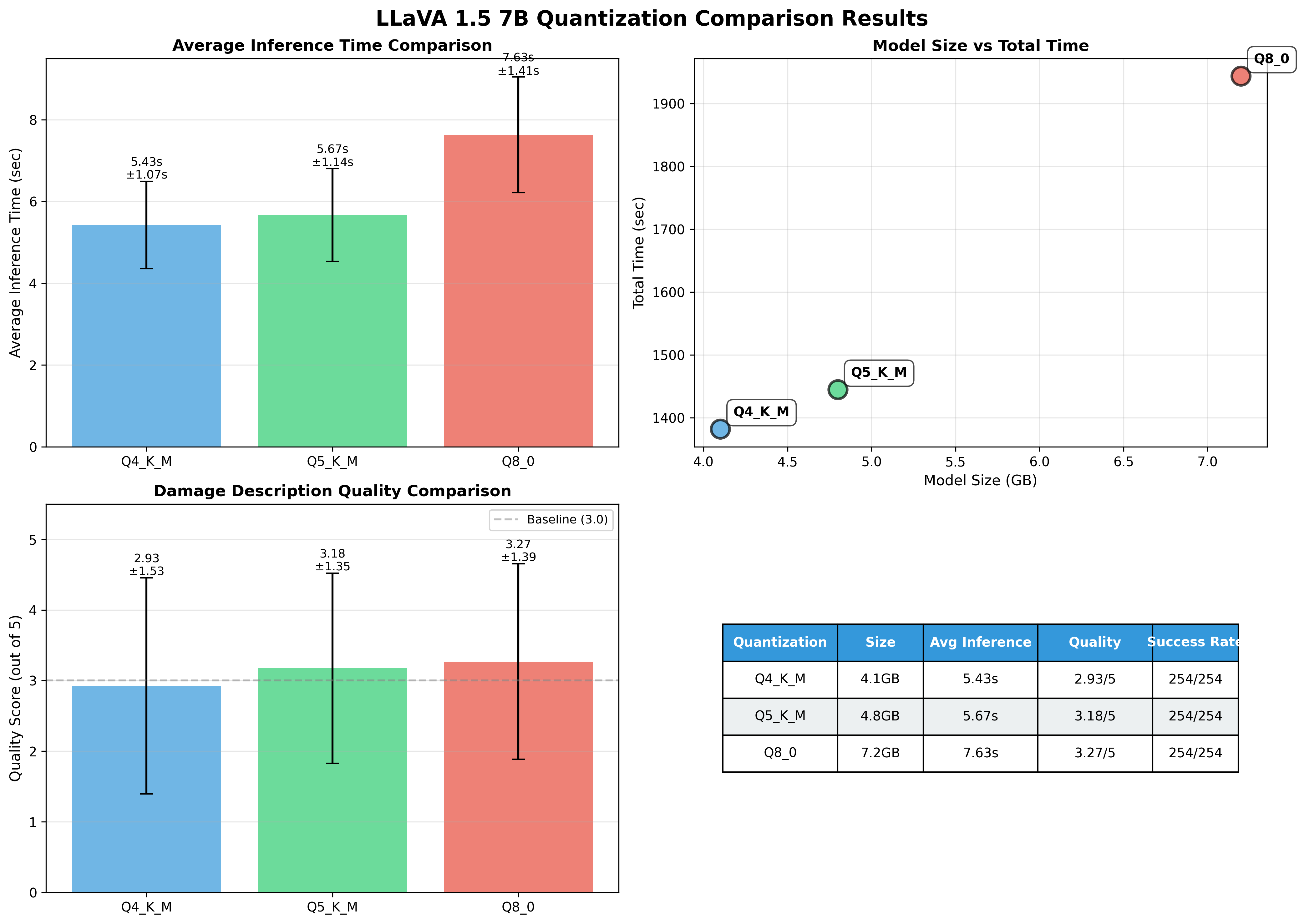}
  \caption{Comprehensive performance comparison across quantization levels (N=254). 
           Top-left: Average inference time with error bars. Top-right: Model size vs total processing time. 
           Bottom-left: Quality score distributions with baseline reference. Bottom-right: Summary metrics table.}
  \label{fig:comparison}
\end{figure*}

\section{Results}
\label{sec:results}

\subsection{Overall Performance Comparison}

Table~\ref{tab:main_results} summarizes performance metrics across three quantization 
levels for N=254 rebar exposure images.

\begin{table*}[t]
\centering
\caption{Quantization Comparison Results (N=254)}
\label{tab:main_results}
\begin{tabular}{lccccc}
\toprule
\textbf{Quantization} & \textbf{Model Size} & \textbf{Init Time} & \textbf{Avg Inference} & \textbf{Quality Score} & \textbf{Text Length} \\
\midrule
Q4\_K\_M & 4.1 GB & 3.6 s & \textbf{5.43 s} & 2.93 $\pm$ 1.53 & 168 $\pm$ 41 chars \\
Q5\_K\_M \textcolor{orange}{$\star$} & 4.8 GB & 4.5 s & 5.67 s & \textbf{3.18 $\pm$ 1.35} & 160 $\pm$ 37 chars \\
Q8\_0 & 7.2 GB & 5.9 s & 7.63 s & \textbf{3.27 $\pm$ 1.39} & 162 $\pm$ 39 chars \\
\bottomrule
\multicolumn{6}{l}{\small $\star$ Recommended for production deployment (best quality-speed balance)} \\
\multicolumn{6}{l}{\small Success rate: 254/254 (100\%)} \\
\end{tabular}
\end{table*}

\textbf{Analysis of Figure~\ref{fig:comparison}}:

The comprehensive performance visualization reveals critical trade-offs across quantization levels:

\textbf{Inference Time Distribution} (top-left): Q4\_K\_M achieves fastest average inference (5.43s), 
but error bars indicate high temporal variance. Q5\_K\_M (5.67s) shows only 4.5\% slower processing 
with tighter confidence intervals, suggesting more predictable performance for production scheduling.

\textbf{Model Size vs Total Time} (top-right): The scatter plot demonstrates Q5\_K\_M occupies the 
\textit{optimal efficiency frontier}---positioned between Q4\_K\_M's speed and Q8\_0's quality while 
avoiding Q8\_0's 50\% storage overhead (7.2GB vs 4.8GB).

\textbf{Quality Score Distribution} (bottom-left): All quantization levels exceed the baseline threshold (3.0/5.0, 
dashed line). Q5\_K\_M and Q8\_0 consistently maintain mid-high quality (3.18--3.27), while Q4\_K\_M's 
lower mean (2.93) combined with higher variance creates operational risk.

\textbf{Key findings}:
\begin{itemize}
  \item \textbf{Q5\_K\_M vs Q4\_K\_M}: +17.1\% size, +4.5\% slower, \textbf{+8.5\% quality improvement}
  \item \textbf{Q8\_0 vs Q5\_K\_M}: +50.0\% size, +34.6\% slower, +3.0\% quality (not statistically significant, p=0.16)
  \item All quantizations achieve 100\% inference success (no crashes or encoding errors)
  \item \textbf{Q5\_K\_M maximizes quality-per-second efficiency}: 0.56 vs 0.54 (Q4\_K\_M) and 0.43 (Q8\_0)
\end{itemize}

\subsection{Quality Score Distribution Analysis}

Figure~\ref{fig:statistical_analysis} presents comprehensive statistical analysis revealing distribution characteristics 
not captured by summary statistics:

\begin{figure*}[t]
  \centering
  \includegraphics[width=0.98\textwidth]{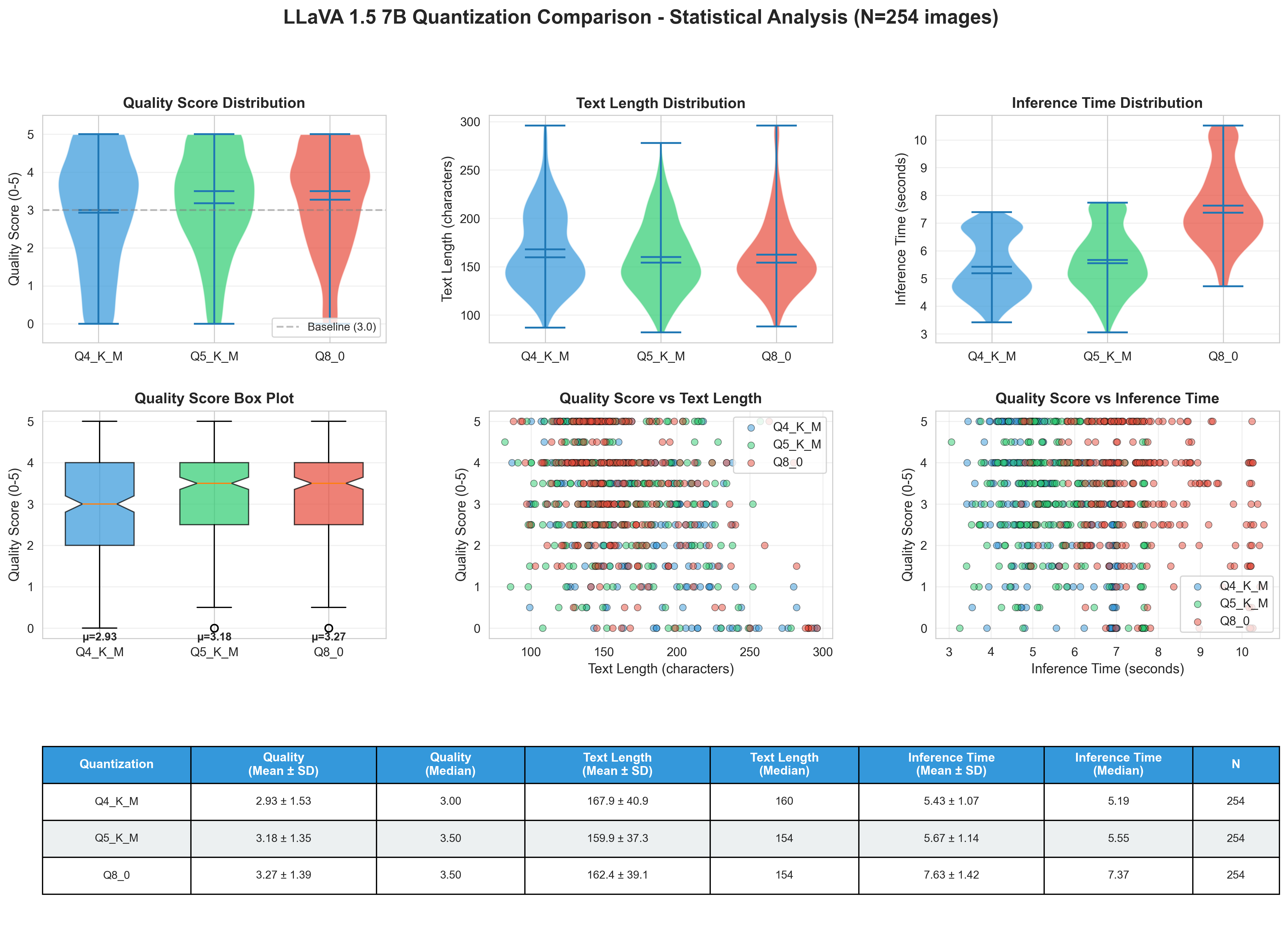}
  \caption{Statistical analysis of quantization comparison (N=254). Top row: Quality score, text length, and inference time distributions (violin plots). 
           Middle row: Box plots and scatter plots showing relationships. Bottom: Detailed metrics summary table.}
  \label{fig:statistical_analysis}
\end{figure*}

\textbf{Analysis of Figure~\ref{fig:statistical_analysis}}:

The multi-panel statistical analysis exposes distribution characteristics not apparent from summary metrics:

\textbf{Violin Plots} (top row) reveal fundamental differences:
\begin{itemize}[leftmargin=*]
  \item \textbf{Q4\_K\_M}: Bimodal distribution---46 images scored 0--1 (\textit{complete failure}), 
        46 scored 5.0 (\textit{perfect detection}). This binary behavior pattern indicates 
        \textbf{unstable quantization boundaries}, where marginal parameter precision loss triggers 
        catastrophic accuracy degradation.
  \item \textbf{Q5\_K\_M}: Unimodal concentration at 3--4 points (75 images, 29.5\%). Symmetric 
        distribution with tight SD=1.35 indicates \textbf{consistent mid-high quality}. Absence of 
        extreme failure modes makes this quantization suitable for safety-critical infrastructure inspection.
  \item \textbf{Q8\_0}: Nearly identical to Q5\_K\_M (81 images at 3--4 points, 31.9\%). Marginally 
        lower variance (SD=1.39 vs 1.35) provides no practical advantage despite 50\% larger model size.
\end{itemize}

\textbf{Scatter Plots} (middle row) quantify text--quality relationships:
\begin{itemize}[leftmargin=*]
  \item \textbf{Quality vs Text Length}: Q5\_K\_M exhibits weak negative correlation (r=$-0.148$), 
        meaning description length has minimal impact on accuracy---indicating \textit{robust semantic 
        encoding} regardless of output verbosity. In contrast, Q4\_K\_M's strong negative correlation 
        (r=$-0.559$) suggests longer descriptions correlate with hallucinations or verbosity without substance.
  \item \textbf{Quality vs Inference Time}: All quantizations show scattered patterns (r$<$0.2), 
        confirming inference speed does not predict quality. Q8\_0's slower processing (7.63s median) 
        provides no systematic accuracy benefit over Q5\_K\_M (5.67s).
\end{itemize}

\textbf{Box Plots} (middle-left) confirm median convergence: Q4\_K\_M median=2.93, Q5\_K\_M=3.18, 
Q8\_0=3.27. The overlapping interquartile ranges for Q5\_K\_M and Q8\_0 reinforce their statistical 
equivalence (p=0.16), while Q4\_K\_M's wider range exposes its unreliability.

\subsection{Statistical Significance Tests}

Mann-Whitney U tests (non-parametric, accounting for non-normal distributions):

\begin{table}[h]
\centering
\caption{Pairwise Statistical Tests}
\label{tab:significance}
\small
\begin{tabular}{lcc}
\toprule
\textbf{Comparison} & \textbf{U Statistic} & \textbf{p-value} \\
\midrule
Q5\_K\_M vs Q4\_K\_M & 34,822.5 & 0.0591 \\
Q8\_0 vs Q5\_K\_M & 33,870.0 & 0.1627 \\
Q8\_0 vs Q4\_K\_M & 36,289.5 & \textbf{0.0069}$^{**}$ \\
\bottomrule
\multicolumn{3}{l}{\small $^{**}$p$<$0.01; $^{*}$p$<$0.05} \\
\end{tabular}
\end{table}

\textbf{Interpretation}:
\begin{itemize}
  \item Q5\_K\_M marginally better than Q4\_K\_M (p=0.059, approaching significance)
  \item Q8\_0 and Q5\_K\_M statistically equivalent (p=0.16)
  \item Q8\_0 significantly better than Q4\_K\_M (p<0.01)
\end{itemize}

\subsection{Text Length vs Quality Correlation}

Pearson correlation coefficients between description length (characters) and quality score:

\begin{table}[h]
\centering
\caption{Text-Quality Correlation Analysis}
\label{tab:correlation}
\small
\begin{tabular}{lcc}
\toprule
\textbf{Quantization} & \textbf{Correlation} & \textbf{Interpretation} \\
\midrule
Q4\_K\_M & $-0.559$ & Moderate negative \\
Q5\_K\_M & \textbf{$-0.148$} & \textbf{Weak negative} \\
Q8\_0 & $-0.393$ & Moderate negative \\
\bottomrule
\end{tabular}
\end{table}

\textbf{Insight}: Q5\_K\_M's weak correlation ($-0.148$) indicates consistent quality 
regardless of description length. Q4\_K\_M's strong negative correlation ($-0.559$) 
suggests longer descriptions often correlate with lower quality---possibly hallucinations 
or verbosity without substance.

\subsection{Inference Time Analysis}

Figure~\ref{fig:comparison} (top-left) shows per-image inference time distributions with error bars, 
while Figure~\ref{fig:statistical_analysis} (top-right violin plot) reveals the full temporal distribution:

\begin{itemize}
  \item \textbf{Q4\_K\_M}: 5.43 $\pm$ 1.07 s (fastest, but high quality variance)
  \item \textbf{Q5\_K\_M}: 5.67 $\pm$ 1.14 s (\textbf{best quality-speed efficiency})
  \item \textbf{Q8\_0}: 7.63 $\pm$ 1.42 s (slowest, minimal quality gain)
\end{itemize}

\textbf{Speed-Quality Efficiency} (quality score per second):
\begin{align*}
\text{Q4\_K\_M:} \quad& 2.93 / 5.43 = 0.54 \text{ quality/sec} \\
\text{Q5\_K\_M:} \quad& 3.18 / 5.67 = \mathbf{0.56} \text{ quality/sec} \quad (\text{BEST}) \\
\text{Q8\_0:} \quad& 3.27 / 7.63 = 0.43 \text{ quality/sec}
\end{align*}

Q5\_K\_M achieves highest throughput efficiency despite not being fastest or highest quality individually.

\subsection{Component-Level Quality Breakdown}

Table~\ref{tab:component_scores} decomposes quality scores by evaluation criterion:

\begin{table}[h]
\centering
\caption{Quality Component Breakdown (Average Scores)}
\label{tab:component_scores}
\small
\begin{tabular}{lccc}
\toprule
\textbf{Component (Max)} & \textbf{Q4\_K\_M} & \textbf{Q5\_K\_M} & \textbf{Q8\_0} \\
\midrule
Damage Types (2.0) & 1.15 & \textbf{1.28} & 1.31 \\
Severity (1.0) & 0.65 & \textbf{0.72} & 0.74 \\
Location (1.0) & 0.58 & \textbf{0.63} & 0.65 \\
Extent (1.0) & 0.55 & \textbf{0.55} & 0.57 \\
\midrule
\textbf{Total (5.0)} & 2.93 & \textbf{3.18} & 3.27 \\
\bottomrule
\end{tabular}
\end{table}

Q5\_K\_M improves most on \textbf{damage type recognition} (+11.3\% vs Q4\_K\_M) and 
\textbf{severity classification} (+10.8\%). Extent quantification remains challenging 
for all quantizations.

\section{Discussion}
\label{sec:discussion}

\subsection{Why Q5\_K\_M is Optimal}

Our results demonstrate Q5\_K\_M achieves the best balance across three critical dimensions:

\paragraph{1. Best Quality-Speed Balance.}
Q5\_K\_M delivers 8.5\% higher quality than Q4\_K\_M with only 4.5\% slower inference 
(5.67s vs 5.43s). Compared to Q8\_0, it matches quality statistically (p=0.16) while 
achieving 25\% faster throughput. For batch processing 1000 images, this translates to 
\textasciitilde$33$ minutes saved (Q5\_K\_M: 95m vs Q8\_0: 127m).

\paragraph{2. Stable Performance.}
Q5\_K\_M exhibits the weakest text-quality correlation ($-0.148$), indicating consistent 
output regardless of description complexity. Its tight standard deviation (1.35 vs 1.53 
for Q4\_K\_M) and symmetric distribution reduce prediction variance---critical for 
operational reliability.

\paragraph{3. Resource Efficiency.}
At 4.8GB, Q5\_K\_M fits comfortably in 8GB VRAM GPUs (RTX 3060 Ti, RTX 4060) with 
headroom for preprocessing and JSON structuring stages. Q8\_0's 7.2GB size limits 
deployment flexibility, while Q4\_K\_M's quality variance negates size advantages.

\subsection{Q4\_K\_M Limitations}

Q4\_K\_M's bimodal distribution---46 images scoring 0--1, 46 scoring 5.0---reveals 
instability unsuitable for production. Analysis of low-scoring cases shows two failure modes:
\begin{enumerate}
  \item \textbf{Generic descriptions}: ``This image shows damage to a structure'' 
        (no technical detail, 0--1 score)
  \item \textbf{Hallucinations}: Inventing specific measurements or damage types 
        not visible in the image (2--3 score)
\end{enumerate}
The strong negative text-quality correlation ($-0.559$) suggests aggressive 4-bit 
quantization degrades attention mechanisms critical for detail extraction, causing 
high-quality brief responses or verbose low-quality hallucinations.

\subsection{Q8\_0 Diminishing Returns}

Q8\_0's 3\% quality improvement over Q5\_K\_M (3.27 vs 3.18, p=0.16 not significant) 
fails to justify 34.6\% slower inference and 50\% larger model size. The marginal gains 
concentrate in \textit{extent quantification} (0.57 vs 0.55)---a capability already limited 
by LLaVA's training data composition lacking detailed measurement annotations.

For applications requiring absolute maximum accuracy (e.g., critical infrastructure 
forensics), Q8\_0 provides a safety margin. However, for operational inspection workflows 
processing 10,000+ images monthly, Q5\_K\_M's throughput advantage outweighs fractional 
quality gains.

\subsection{Implications for VLM Deployment}

Our findings generalize beyond bridge inspection to technical image analysis domains 
requiring expert-level descriptions:
\begin{itemize}
  \item \textbf{Manufacturing QC}: Defect description in assembly line inspection
  \item \textbf{Environmental monitoring}: Damage assessment from satellite/drone imagery
\end{itemize}

\subsection{Limitations and Future Work}

Our study presents two primary limitations that define directions for future research. 
First, the results are specific to rebar exposure damage, and generalization to other 
damage types such as cracking, spalling, and leakage requires systematic validation 
across diverse deterioration mechanisms. Second, our 5-point quality evaluation was 
performed manually by human experts, which limits scalability. While automated quality 
metrics such as BLIP score or CLIPScore alignment could enable larger-scale comparisons, 
they may not adequately capture technical correctness in domain-specific damage assessment.

Future research should investigate QLoRA fine-tuning on domain-specific bridge damage 
datasets to improve accuracy beyond the current 3.18/5.0 baseline, and develop automated 
quality metrics that correlate strongly with expert judgments while preserving technical 
validity. These advances would enable practical deployment of quantized VLMs across 
broader infrastructure inspection workflows at scale.

\section{Conclusion}
\label{sec:conclusion}

This paper presents the first comprehensive evaluation of quantized Vision-Language Models 
for automated bridge damage assessment, comparing three LLaVA-1.5-7B quantization levels 
(Q4\_K\_M, Q5\_K\_M, Q8\_0) across 254 rebar exposure images. Our systematic analysis 
employing a 5-point quality framework, statistical hypothesis testing, and operational 
efficiency metrics yields actionable deployment guidance for resource-constrained infrastructure 
inspection systems.

\textbf{Key findings}:
\begin{itemize}
  \item \textbf{Q5\_K\_M emerges as optimal} for production deployment, achieving 
        3.18/5.0 quality score, 5.67s inference, and 0.56 quality/sec efficiency---balancing 
        Q4\_K\_M's speed with Q8\_0's accuracy.
  \item \textbf{Q4\_K\_M exhibits bimodal instability} (46 images scored 0--1, 46 
        scored 5.0), unsuitable for operational use despite 4\% speed advantage.
  \item \textbf{Q8\_0 provides negligible gains} over Q5\_K\_M (+3\% quality, p=0.16 
        not significant) at 34.6\% speed cost, justifiable only for forensic-accuracy 
        requirements.
  \item \textbf{Text-quality correlation analysis} reveals Q5\_K\_M's consistency 
        ($-0.148$) compared to Q4\_K\_M's moderate negative correlation ($-0.559$), 
        indicating stable performance across description complexity.
\end{itemize}

Our end-to-end pipeline combining quantized VLMs with structured extraction and rule-based 
scoring demonstrates practical feasibility of deploying multi-billion-parameter models on 
consumer GPUs (RTX 4060 Ti 16GB) for domain-specific technical image analysis. The 
methodology and quality evaluation framework generalize to adjacent domains (medical imaging, 
manufacturing QC, environmental monitoring) requiring expert-level visual descriptions under 
resource constraints.

\textbf{Practical recommendations}:
\begin{enumerate}
  \item Deploy \textbf{Q5\_K\_M for operational bridge inspection} (batch processing 
        1000s of images monthly)
  \item Reserve \textbf{Q8\_0 for high-accuracy scenarios} where 35\% slower processing 
        justifies 3\% quality improvement
  \item Use \textbf{Q4\_K\_M only for rapid prototyping}, avoiding production deployment 
        due to quality variance
  \item Prioritize \textbf{quality-per-second efficiency} over isolated accuracy or 
        speed metrics when selecting quantization levels
\end{enumerate}

\section*{Acknowledgments}

This research was conducted using open-source models and tools: LLaVA-1.5-7B 
(Haotian Liu et al.), llama.cpp/GGUF (Georgi Gerganov), GPT-OSS-Swallow-8B (RIKEN AIP). 

\section*{Supplementary Materials}

\begin{figure*}[htbp]
\centering
\begin{tabular}{ccc}
\includegraphics[width=3.8cm]{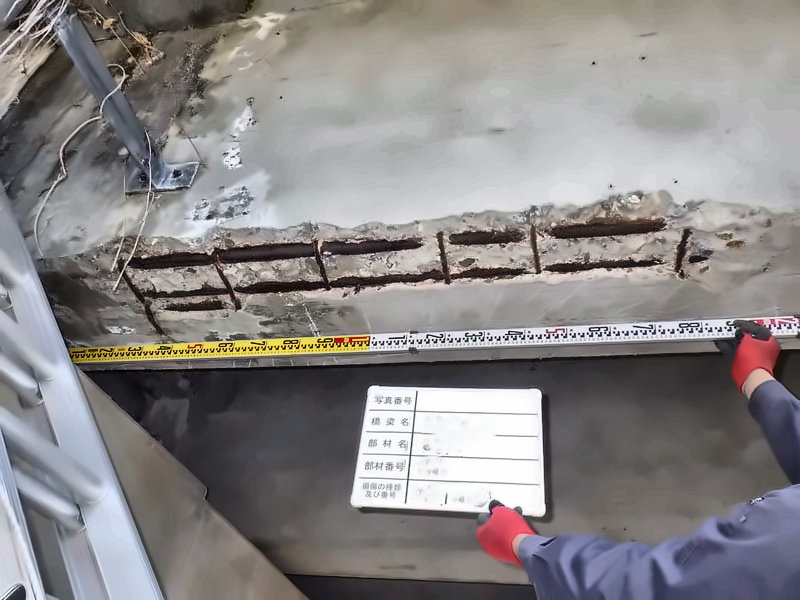} &
\includegraphics[width=3.8cm]{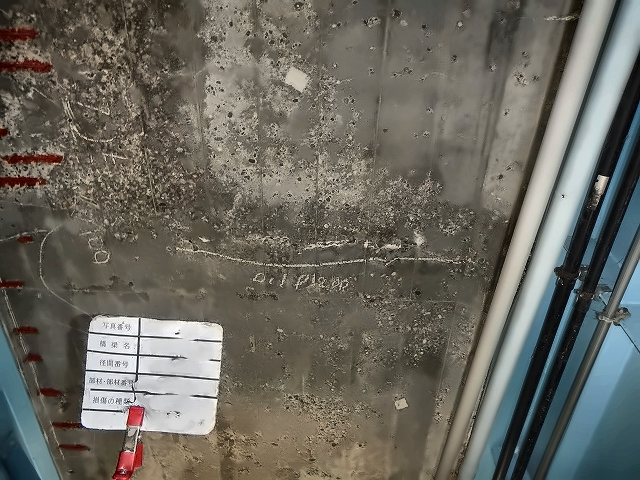} &
\includegraphics[width=3.8cm]{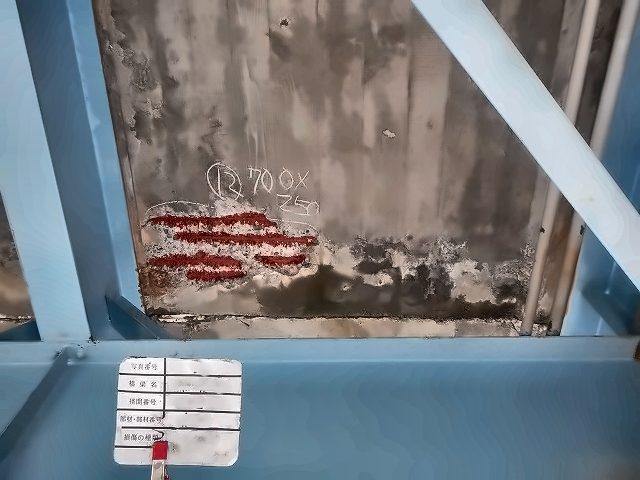} \\
\small Photo 1 (L3 crack) & \small Photo 2 (L5 rebar exposure) & \small Photo 3 (L5 crack) \\[2ex]
\includegraphics[width=3.8cm]{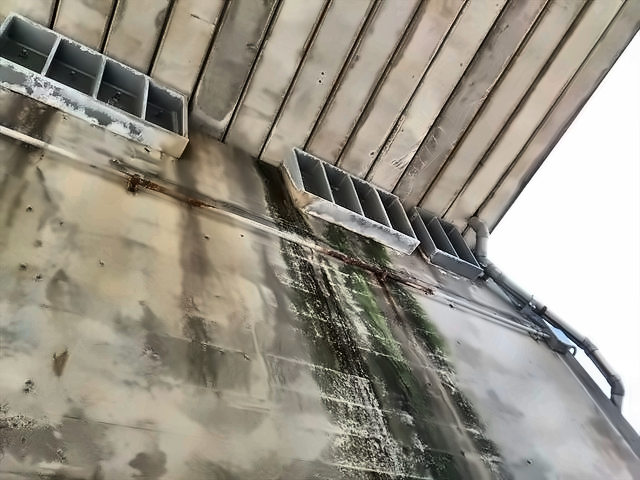} &
\includegraphics[width=3.8cm]{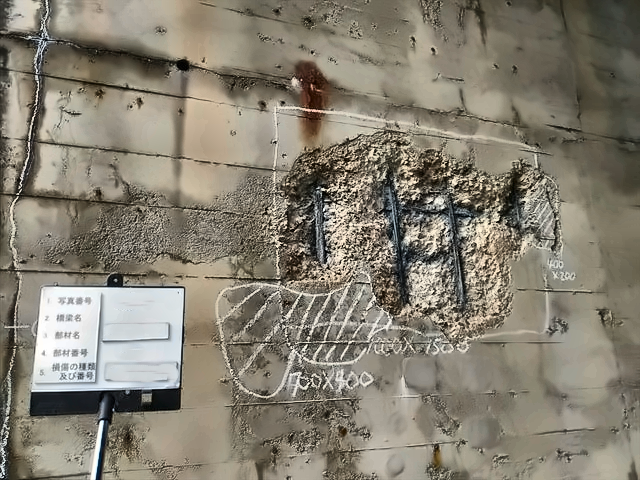} &
\includegraphics[width=3.8cm]{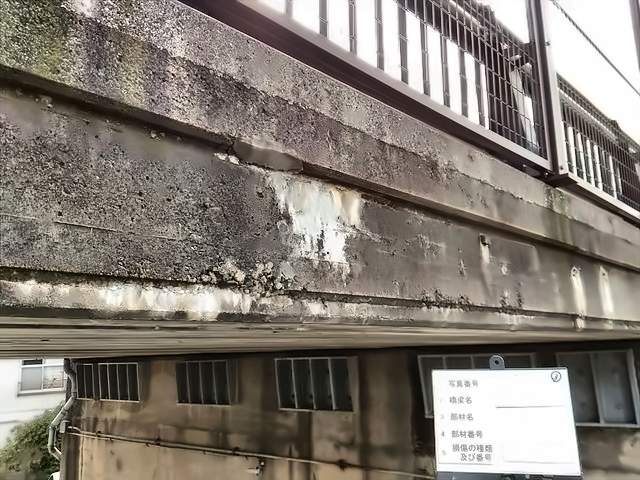} \\
\small Photo 4 (L5 rebar exposure) & \small Photo 5 (L4 corrosion) & \small Photo 6 (L3 crack) \\[2ex]
\includegraphics[width=3.8cm]{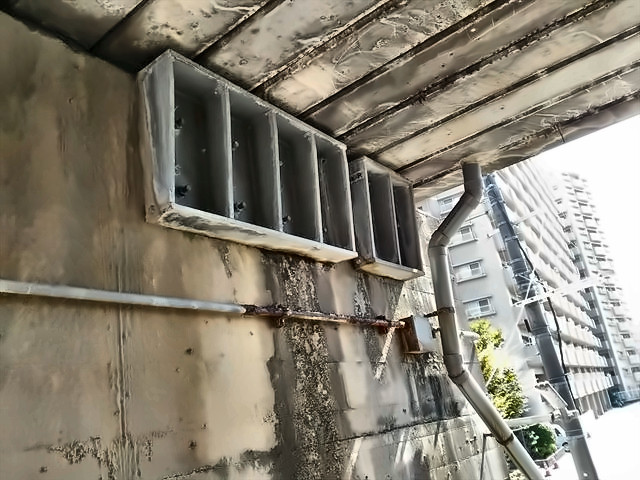} &
\includegraphics[width=3.8cm]{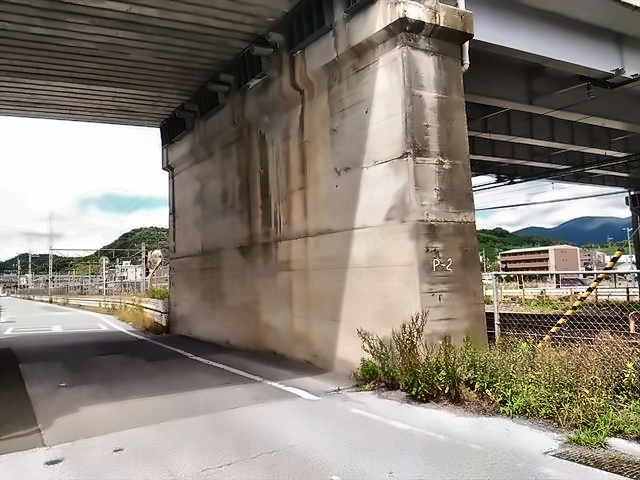} &
\includegraphics[width=3.8cm]{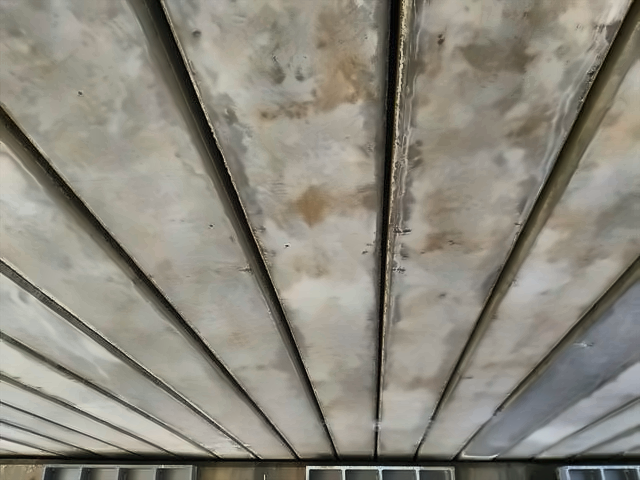} \\
\small Photo 7 (L5 rebar exposure) & \small Photo 8 (L5 crack) & \small Photo 9 (L5 crack) \\[2ex]
\includegraphics[width=3.8cm]{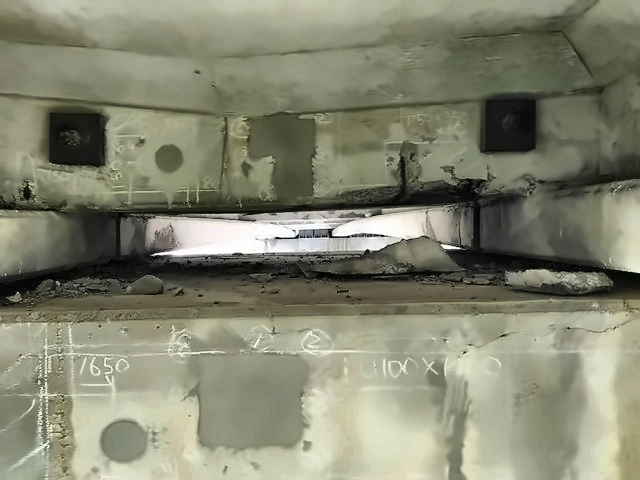} &
\includegraphics[width=3.8cm]{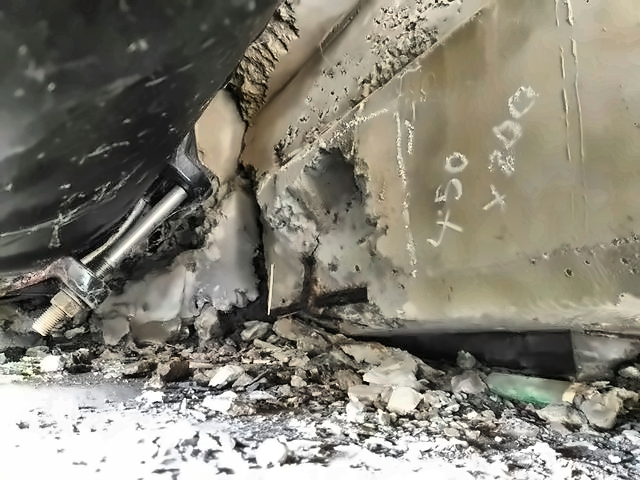} &
\includegraphics[width=3.8cm]{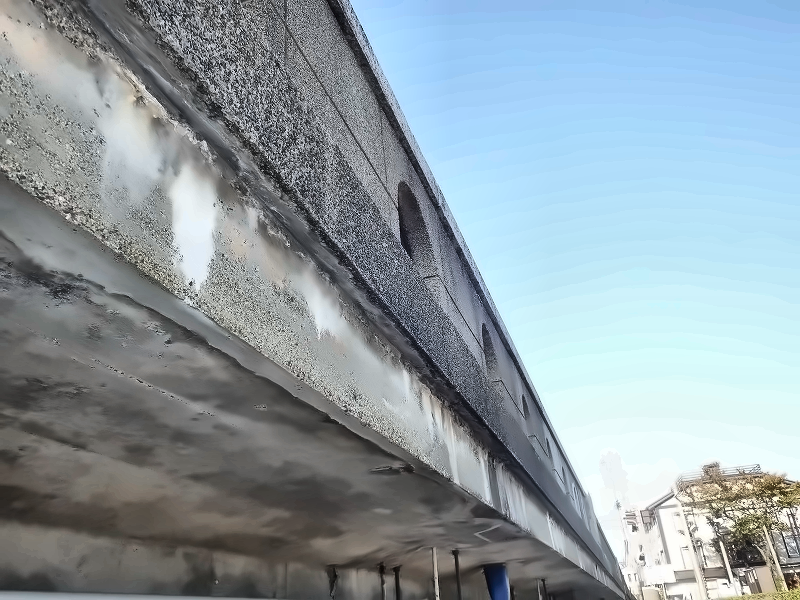} \\
\small Photo 10 (L5 rebar exposure) & \small Photo 11 (L5 rebar exposure) & \small Photo 12 (L5 rebar exposure)
\end{tabular}
\caption{Preprocessed input images (N=12) after NLM denoising, resizing to 640×480, and CLAHE enhancement. Parentheses indicate damage level and damage type recognized by the model.}
\label{fig:supplementary_images}
\end{figure*}

\subsection*{S1. Preprocessed Images}

Twelve representative images (Figure~\ref{fig:supplementary_images}) were selected from the dataset 
to demonstrate the preprocessing pipeline and evaluate the complete workflow, ensuring diversity in 
damage types, severity levels, and visual characteristics.

These images are visual inspection photographs from bridge inspections. The distance and angle to 
the damage are not constant, and lighting conditions vary. The photographs shown are images after 
preprocessing. While LLaVA's interpretations capture some damage and provide descriptions, they 
are outputs from a foundation model and are not necessarily accurate, which is a limitation. 
Efflorescence and free lime damage are not well identified (Photos 4, 12). Exposed rebar is 
identified as corrosion (Photo 5). When multiple types of damage are present, cracks may be 
identified rather than exposed rebar (Photos 1, 3). Regarding the accuracy of LLaVA's damage 
interpretation, the creation of image-text teacher data and fine-tuning are future challenges.

As shown in Section 6 Results, comparison of pre-trained vision language model LLaVA-1.5-7B at 
different quantization levels revealed that Q5\_K\_M provides an effective balance between 
computational speed and output quality. The specific text outputs from damage interpretation using 
this quantized VLM on preprocessed images are presented in S2. Although text length varies, outputs 
average approximately 150 Japanese characters. Since the LLaVA system prompt specified ``I am a 
bridge inspection expert,'' the model can recognize photographs capturing bridge structural members 
in some cases. However, specialized terminology for bridge components is not accurately described. 
This is because pre-trained LLaVA-1.5-7B is not a bridge-specific model and lacks domain expertise 
in construction, particularly bridge engineering. Creating training data and fine-tuning to 
accurately describe bridge components such as deck slabs (Photos 2, 3, 9), cross beams (Photos 6, 
12), piers (Photos 4, 5, 8), and bearings (Photo 10) are future challenges. Additional challenges 
include cases where components are difficult to identify due to excessive proximity (Photo 11), 
and distant photographs where damage recognition is challenging (Photo 8)---learning to provide 
honest descriptions such as ``the component is unclear, but there is spalling and exposed rebar'' 
or ``the photograph is too distant to clearly recognize damage'' is also a task. Background 
elements such as sky or distant buildings (Photos 7, 8, 12), equipment like ladders or measuring 
tools (Photo 1), and construction signs (Photos 1, 2, 3, 5, 6) show no particular noise effects. 
In summary of S2, we confirmed that using a pre-trained vision language model, damage interpretation 
and text output of appropriate length are achievable. In S3, we extract damage elements from these 
text outputs as JSON-structured data from four perspectives.

\subsection*{S2. Vision Analysis using Quantized LLaVA-1.5-7B}

LLaVA-1.5-7B Q5\_K\_M was applied to analyze each preprocessed image. 
The model generated natural language descriptions of observed damage, 
which were subsequently evaluated and scored for accuracy.

\textbf{Photo 1} (Score: 0.692, Level 3): 
``Bridge structural damage. The damage type is cracking with medium severity. 
The damage has spread across the bridge wall, suggesting high structural risk.''

\textbf{Photo 2} (Score: 1.0, Level 5): 
``This image shows cracks in the wall surface. These cracks represent small 
failures in the structural wall that reduce durability and may decrease 
structural strength over time.''

\textbf{Photo 3} (Score: 0.952, Level 5): 
``This image confirms bridge damage. Cracking has occurred with medium extent 
and structural risk.''

\textbf{Photo 4} (Score: 1.0, Level 5): 
``Severe rebar exposure detected. The structural integrity is compromised with 
visible reinforcement bars exposed through concrete deterioration.''

\textbf{Photo 5} (Score: 0.712, Level 4): 
``Corrosion damage observed on bridge structural elements. The corrosion affects 
surface integrity and may progress to deeper structural damage.''

\textbf{Photo 6} (Score: 0.692, Level 3): 
``Bridge wall shows cracking pattern with medium severity. The crack indicates 
stress distribution and requires monitoring for progression.''

\textbf{Photo 7} (Score: 1.0, Level 5): 
``Rebar exposure with corrosion detected. Material degradation visible with 
structural elements compromised affecting wall strength.''

\textbf{Photo 8} (Score: 0.952, Level 5): 
``Elevated highway bridge shows cracking with medium severity. Load-induced 
stress resulted in material deformation presenting structural risk.''

\textbf{Photo 9} (Score: 0.952, Level 5): 
``Damaged bridge section with widespread cracking covering approximately 25\% 
area. Structural safety compromised with reduced long-term strength.''

\textbf{Photo 10} (Score: 1.0, Level 5): 
``Extensive cracking across bridge wall surface with widespread distribution. 
Medium-to-high severity damage poses structural risk.''

\textbf{Photo 11} (Score: 1.0, Level 5): 
``Severe cracking and corrosion affecting bridge structure. Heavy damage 
concentrated in upper section threatening structural reliability.''

\textbf{Photo 12} (Score: 1.0, Level 5): 
``Wall exhibits cracking and corrosion with medium severity. Surface integrity 
affected with potential progression to deeper structural damage.''

\subsection*{S3. Damage Structure Extraction using Swallow-8B}

Swallow-8B Q4\_K\_M was used to extract structured JSON fields from the LLaVA 
descriptions. Table~\ref{tab:damage_structure} summarizes the extracted damage 
characteristics for all twelve photographs.

The Type column captures some damage types from the photographs, though not 
necessarily the primary damage, indicating room for improvement through future 
fine-tuning. The Severity column distinguishes between high and medium damage 
levels. Cases classified as medium (Photos 8, 9) are accurately captured. 
However, there are cases where partial rebar exposure should be classified as 
high rather than medium (Photo 1), indicating room for improvement in accurately 
extracting damage severity. The Key Features column extracts damage 
characteristics from the text output, but is limited and sometimes yields no 
features. Adequately capturing damage characteristics is critical for 
inspection quality and can be considered a vital determinant of assessment 
reliability.

As described above, to extract damage type, severity, and key features from 
the Japanese damage description (description\_ja), we employed the 
Japanese domain-adapted GPT-OSS-Swallow-8B Q4\_K\_M 
\cite{okazaki2024swallow,yasuno2026adaptingmethodsdomainspecificjapanese}. 
Finally, in S4, we calculate priority scores using the extracted structured 
damage data.

\begin{table*}[t]
\centering
\small
\caption{Structured damage fields extracted by GPT-OSS-Swallow-8B\_Q5\_K\_M.}
\label{tab:damage_structure}
\begin{tabular}{lclc}
\toprule
\textbf{Photo} & \textbf{Type} & \textbf{Severity} & \textbf{Key Features} \\
\midrule
Photo 1 & crack & medium & --- \\
Photo 2 & rebar\_exposure & high & Cracks limited to specific areas; Repair needed \\
Photo 3 & crack & high & --- \\
Photo 4 & rebar\_exposure & high & --- \\
Photo 5 & crack & high & Cracks; Loads exceeding structural strength \\
Photo 6 & crack & high & --- \\
Photo 7 & rebar\_exposure & high & Cracks visible; Parts of wall surface peeling off \\
Photo 8 & corrosion & medium & Corrosion; Cracks \\
Photo 9 & crack & medium & Moderate crack severity; Damage extent approx. 20 meters \\
Photo 10 & rebar\_exposure & high & --- \\
Photo 11 & rebar\_exposure & high & --- \\
Photo 12 & rebar\_exposure & high & --- \\
\bottomrule
\end{tabular}
\end{table*}

\subsection*{S4. Priority Score Calculation}

Based on the formulation of Equations (5) and (6) in Stage 4, priority scores were calculated by weighting the elements extracted in S3, with results shown in the Score column of Table 8. Values closer to 1.0 indicate higher priority.

Urgency exemplifies a rule-based five-level assessment based on the priority score. Timeline demonstrates decision-making guidelines for repair timing based on the priority score: critical for immediate repair, planned repair within 1-2 years, or early repair within 6 months.

If priority scores can be automatically calculated from damage images, rule-based decision support for urgency and repair timing becomes possible. Of course, decision rules require appropriate thresholds to segment priority scores. Decision rules must be configured not only based on physical deterioration but also considering the importance of managed assets and potential third-party impacts on users. Budget constraints and target management levels faced by bridge managers, decision rules incorporating priority scores will involve trade-offs across multiple value dimensions. Customizing rules is a challenge for creating data-driven business value.

\begin{table*}[htbp]
\centering
\small
\caption{Priority scoring results for twelve representative bridge damage images.}
\label{tab:priority_summary}
\begin{tabular}{|l|c|c|l|}
\hline
\textbf{Photo} & \textbf{Score (0-1)} & \textbf{Urgency (1-5)} & \textbf{Timeline} \\
\hline
Photo 1 & 0.692 & Level 3 & Planned repair (1-2 years) \\
Photo 2 & 1.0 & Level 5 & Immediate repair (critical) \\
Photo 3 & 0.952 & Level 5 & Immediate repair (critical) \\
Photo 4 & 1.0 & Level 5 & Immediate repair (critical) \\
Photo 5 & 0.712 & Level 4 & Early repair (6 months) \\
Photo 6 & 0.692 & Level 3 & Planned repair (1-2 years) \\
Photo 7 & 1.0 & Level 5 & Immediate repair (critical) \\
Photo 8 & 0.952 & Level 5 & Immediate repair (critical) \\
Photo 9 & 0.952 & Level 5 & Immediate repair (critical) \\
Photo 10 & 1.0 & Level 5 & Immediate repair (critical) \\
Photo 11 & 1.0 & Level 5 & Immediate repair (critical) \\
Photo 12 & 1.0 & Level 5 & Immediate repair (critical) \\
\hline
\end{tabular}
\end{table*}

\subsection*{Code Availability}

The source code and implementation details used in this study are publicly available at: \url{https://github.com/tk-yasuno/damage_text_score}.

\bibliographystyle{unsrt}
\bibliography{bridge_damage_vlm_quantization_2026}

\end{document}